%% file: main.tex
\documentclass[conference]{IEEEtran}
\IEEEoverridecommandlockouts
\def\IEEEtitletopspaceextra{6mm}

\usepackage{cite}
\usepackage{amsmath,amssymb}
\usepackage{algorithm}
\usepackage{algorithmic}
\usepackage{array}
\usepackage{booktabs}
\usepackage{graphicx}
\usepackage{textcomp}
\usepackage{tikz}
\usetikzlibrary{arrows.meta,positioning,calc}
\usepackage{longtable}
\newcolumntype{P}[1]{>{\raggedright\arraybackslash}p{#1}}
\usepackage[hidelinks,
  pdftitle={Auditing Targetless Camera--LiDAR Extrinsic Correction before Recursive Use},
  pdfauthor={Bo Ma, Weiqi Yan}]{hyperref}

\title{Auditing Targetless Camera--LiDAR Extrinsic Correction before Recursive Use}

\author{\IEEEauthorblockN{Bo Ma and Weiqi Yan}
\IEEEauthorblockA{Auckland University of Technology, Auckland 1024, New Zealand\\
rcn4743@aut.ac.nz, weiqi.yan@aut.ac.nz}}

\begin{document}

\maketitle

\begin{abstract}
Camera--LiDAR fusion repeatedly uses an extrinsic transform in projection,
association, and recursive state estimation. We study whether a targetless
correction should be admitted to a downstream consumer only after separate
audits of correction quality, confidence ranking, and recursive risk. Our
log-disjoint protocol evaluates these stages under synthetic extrinsic drift
on Argoverse~2 and treats cross-dataset and odometry experiments as bounded
diagnostics.

On 27 test-split logs, single-frame Canny alignment is harmful in the mild
tier. Multi-frame alignment improves severe-tier rotation by $2.52^\circ$
but worsens translation by $0.177$\,m; its declared joint error-magnitude
improvement is also negative ($-0.158$, 95\% CI $[-0.223,-0.098]$), so the
rotation gain is not a six-degree-of-freedom correction success. A
zero-injection check moves away from the true extrinsic, and solver-disagreement
AUROC on selected splits does not yield a validated nonzero-coverage
risk--coverage operating point.

In a 15-log final system role, the locked conservative policy admits no
candidate and equals Raw, whereas Always-correct increases trajectory error.
These results support an auditable separation of correction, confidence, and
recursive contamination; they do not establish a deployable calibration gate,
positive odometry or SLAM improvement, map consistency, or closed-loop safety.
\end{abstract}

\section{Introduction}
\label{sec:intro}

Camera--LiDAR fusion systems increasingly consume LiDAR point clouds
projected into the camera frame, or bird's-eye-view features fused from
both modalities, as evidence for 3D scene understanding
\cite{liu2023bevfusion}.
Every such pipeline assumes a particular extrinsic transform $T$ between
the LiDAR and camera frames is correct. In practice $T$ is estimated
once, offline, and treated as a constant; real vehicles experience
rotation and translation drift around that estimate from vibration,
thermal cycling, and mounting tolerances. Temporal misalignment is a
separate failure mode and is outside the evaluated protocol. A
system that silently trusts a stale $T$ propagates the resulting
misprojection into every cross-modal judgment it makes, with no signal
that this happened.

Two adjacent literatures address pieces of this problem without
addressing it jointly. Learning-based extrinsic calibration
\cite{schneider2017regnet,iyer2018calibnet,lv2021lccnet} regresses $T$
(or a correction to it) from paired image/point-cloud evidence, and
recent work adds calibrated uncertainty via conformal prediction
\cite{cocheteux2025uncertainty}; this line reports interval coverage on
the calibration error itself, not the downstream effect on a 3D
understanding task. Learned driving-perception systems such as BEVFusion
\cite{liu2023bevfusion} consume calibrated sensor geometry as a given input
and are not evaluated here under deliberately injected extrinsic drift. We ask the bounded
operational question: \emph{when a downstream 3D understanding system is
given a possibly miscalibrated extrinsic, can it detect and recover from it
without trusting a separate offline calibration procedure blindly?}

This paper makes a narrower claim. We introduce a confidence-gated protocol
for extrinsic correction and test whether its components can be made reliable
under synthetic drift. The
single-frame edge-alignment instantiation fails under mild drift, while
multi-frame evidence recovers a severe-tier rotation signal and descriptive
confidence ranking but does not establish a safe two-tier gate. Sections~\ref{sec:method}--\ref{sec:eval}
define the protocol and its stopping rules; Sections~\ref{sec:status} and
\ref{sec:remediation} report the observed evidence. A recursive
visual--LiDAR odometry extension is evaluated as a bounded negative result, not as
a claim of improved odometry or deployment safety.

\textbf{Scope.} We study whether a \emph{guarded correction module} can
correct synthetic extrinsic drift without applying harmful corrections
when its confidence is unreliable. We do not claim a general-purpose
calibration-quality certificate -- a caller who wants only a calibration
error bound should use a dedicated audit such as
\cite{cocheteux2025uncertainty} -- and we do not claim positive
bounded visual--LiDAR odometry, trajectory, map, or closed-loop control performance. A stage whose
admission criteria are not met is reported as negative with its criteria
and observed values, not dropped or re-run with relaxed criteria.

\section{Related Work}
\label{sec:related}

\textbf{Calibration.} Targetless calibration aligns projected LiDAR
discontinuities with image structure \cite{levinson2013automatic}; learned
methods regress or geometrically supervise the 6-DoF transform
\cite{schneider2017regnet,iyer2018calibnet,lv2021lccnet}. Recent work adds
conformal uncertainty for the calibration error itself
\cite{cocheteux2025uncertainty}, but does not audit whether a correction is
safe for a downstream recursive estimator. Our contribution is a log-disjoint
reliability audit and negative case study that exposes this missing
correction--confidence--consumer separation.

\textbf{Recursive fusion and uncertainty.} Visual--LiDAR odometry uses the
extrinsic in feature association and motion estimation
\cite{graeter2018limo}; this is the recursive consumer
audited here.
BEVFusion \cite{liu2023bevfusion} represents the downstream fusion setting.
Conformal prediction
\cite{cocheteux2025uncertainty} motivates auditing a score
before using it as a gate; solver disagreement is an analogous uncertainty
proxy, but here it measures a classical cost surface rather than model
parameter uncertainty.

\section{Method}
\label{sec:method}

\textbf{Problem statement.} Both $T$ and $\hat T$ map camera coordinates to
the ego frame. Given image $I$, point set ${\cal P}$, and a possibly drifted
extrinsic $\hat T$, Stage~1 estimates a 6-DoF correction $\Delta\hat T$ and
confidence. We define $T\ominus\hat T=\hat T^{-1}T$ in the camera-local
right-action convention, so $T=\hat T\Delta T$ in homogeneous coordinates.
With $J$ denoting an image--LiDAR alignment cost and ${\cal B}$ a fixed
search box,
\begin{equation}
  \begin{aligned}
  \Delta\hat{T}&=\arg\min_{\Delta T\in{\cal B}}
  J(\Delta T;I,{\cal P},\hat{T}),\\
  T_{\mathrm{corr}}&=\hat{T}\Delta\hat{T},
  \end{aligned}
  \label{eq:correction}
\end{equation}
For our camera-local right-action convention,
$R_{\rm corr}=\hat R\Delta R$ and
$t_{\rm corr}=\hat t+\hat R\Delta t$. We report rotation geodesic error,
translation error, and the declared, task-scaled composite spatial score
$d_{\mathrm{comp}}(s_t)=\sqrt{\theta_{\rm rad}^2+(\|t_1-t_2\|/s_t)^2}$, with the
primary analysis using $s_t=1\,\mathrm{m}$ and a pre-specified sensitivity audit
at $s_t\in\{0.1,1,10\}\,\mathrm{m}$. This is an error-magnitude reporting
composite, not an invariant geodesic metric on $SE(3)$. The consumer admits a
correction only after multistart and temporal checks; otherwise it retains
$\hat T$.

\begin{figure}[t]
\centering
\begin{tikzpicture}[
  stage/.style={draw, rounded corners, fill=gray!10, minimum width=1.9cm, minimum height=0.85cm, align=center, font=\scriptsize},
  ann/.style={font=\tiny, align=center, text width=1.9cm},
  arrowstyle/.style={-{Latex[length=1.3mm]}, thick}
]
\node[stage] (s1) {Stage 1\\$\Delta T$ + confidence};
\node[stage, right=5mm of s1] (s2) {Stage 2\\temporal guard + odometry};
\draw[arrowstyle] (s1) -- (s2);
\node[ann, below=1mm of s1] {weakly supervised;\\AUROC-audited};
\node[ann, below=1mm of s2] {admit/lock or\\fallback to $\hat{T}$};
\coordinate (g1) at ($(s1.east)!0.5!(s2.west)$);
\draw[dashed, red, thick] (g1) ++(0,0.85) -- ++(0,-1.8)
  node[midway, above, font=\tiny, red, sloped, xshift=-4mm] {gated};
\end{tikzpicture}
\caption{The evaluated two-stage protocol connects a frozen correction
module to a bounded visual--LiDAR odometry consumer through an explicit gate.
The consumer uses a correction only when it is admitted and locked for a
segment; otherwise it falls back to $\hat{T}$. The extension evaluates longer
trajectories and reports reprojection/association proxies when an HD-map
archive is unavailable.}
\label{fig:stages}
\end{figure}
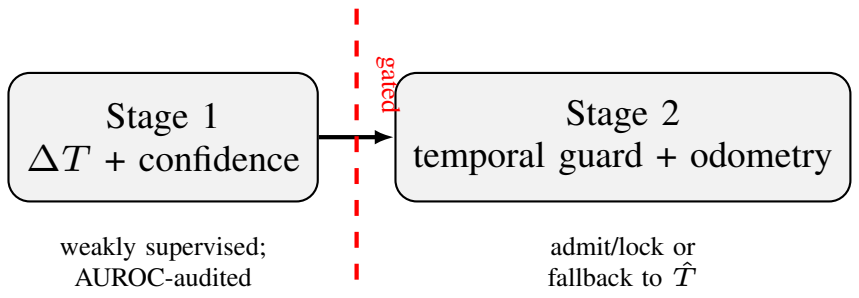

\textbf{Stage 1 (evaluated protocol).} Inputs are the front-center camera image,
the LiDAR point cloud, and a perturbed extrinsic $\hat T$; the true $T$ is
used only to generate the training label $\Delta T=T\ominus\hat T$ and
to evaluate errors, never as a model input. Argoverse~2
\cite{wilson2021argoverse2} supplies three pre-declared tiers: \emph{correct}
($\Delta=0$), \emph{mild} (rotation $0.5^\circ$--$2^\circ$,
translation 2--10\,cm), and \emph{severe} (rotation $5^\circ$--$10^\circ$,
translation 20--35\,cm). These ranges are fixed before evaluation and
cover the spatial error magnitudes used in prior calibration work
\cite{schneider2017regnet,iyer2018calibnet}. The model outputs a 6-DoF
correction and a scalar confidence; it is not given the injected severity.

\textbf{Information boundary.} The true extrinsic is used only for weak-label
construction, offline error evaluation, and the non-deployable Oracle arm;
logged ego pose is used only for offline sweep compensation. Injected tier,
learned targets, and final reference poses are restricted to stratification,
training, and evaluation respectively: none is an online feature, score,
threshold-selection input, or final-rollout decision. Table~\ref{tab:data-boundary}
puts the split roles and privileged-data boundary in one place; extended
parameter tables are given in Appendix~\ref{app:reproducibility}.

\begin{table}[t]
\centering
\scriptsize
\setlength{\tabcolsep}{2pt}
\renewcommand{\arraystretch}{1.05}
\begin{tabular}{@{}p{0.34\linewidth}p{0.62\linewidth}@{}}
\toprule
Data role / fixed sample & Boundary and prohibited inference \\
\midrule
Stage~1 formal: AV2 test, 27 logs; every 3rd frame;
1,435 frames/tier/condition & True $T$: weak labels and offline metrics only;
no test-derived threshold or decision \\
Learned split: same 27 logs, fixed 18/9
train/holdout; 473 holdout frames/tier & Heatmap targets use $T$ on train only;
holdout is not used for tuning \\
Exploratory ranking: AV2 train/val, 60 logs;
post-hoc 30/30 split & Severity is for stratification only; ranking/risk
scans cannot select the system threshold \\
System audit: 3/15/15 development/calibration/final
logs; 100 sweeps, $\geq30$ frames, target 50 & Calibration locks; final is evaluation only;
logged ego pose is offline/oracle compensation \\
External diagnostic: 170/850 nuScenes Trainval
scenes; no locked manifest hash & Plumbing/failure-pattern diagnostic only;
no cross-domain generalization claim \\
\bottomrule
\end{tabular}
\caption{Data roles and privileged-information boundary; no truth enters
online decisions.}
\label{tab:data-boundary}
\end{table}

\textbf{Stage 2.} The recursive consumer compares raw $\hat T$, oracle $T$,
always-correct, and guarded policies on the same fixed-drift logs. All arms
share timestamps, initial state, frame schedule, observations, and seed. The
guard locks only after agreeing independent re-solves and temporally
consistent adjacent windows; a failed current check resets the streak, and
decisions use only current or past observations. Raw and Oracle ignore the
guard schedule, while Always-correct applies the current candidate. The
online-motion extension's locked never-admit outcome remains a negative
recursive-risk result, not positive downstream-odometry or deployment evidence.

The guard is a finite state machine: each window starts in \emph{Rejected}; a
window satisfying both observable checks increments the consecutive-pass
streak, while any failed check resets it to \emph{Rejected}. After $H$
consecutive passes the current candidate enters \emph{Locked} for its declared
segment. No future window is read to make the current decision, and a segment
reset returns to \emph{Rejected}; the four arms use the same schedule and
observations.

For a $K$-frame window, the shared correction minimizes
$J_K(\Delta T)=K^{-1}\sum_kJ_k(\Delta T)$ after each sweep is
motion-compensated into the reference camera frame. The online score
$b_w$ is the mean pairwise rotation-geodesic distance among the canonical
solve and randomized re-solves in window $w$; $a_w$ measures disagreement
with preceding windows. The legacy score is a clipped relative cost drop
multiplied by a point-count factor. These scores are used only for ranking
until a threshold is selected on calibration data. Recorded ego poses are
used for offline sweep-to-camera compensation, so the multi-frame result
does not validate an estimated online motion model. A separate no-motion or
noisy-pose dependency audit is therefore not silently substituted for the
reported matrix.

\section{Evaluation Protocol and Admission Gates}
\label{sec:eval}

\textbf{Statistical hierarchy.} We declare one primary unit, comparison,
safety constraint, and confidence interval before any secondary analysis:
the log (log-disjoint splits, log-level resampling); Guarded versus Raw
tier-blind trajectory error on the 15-log final split
(Table~\ref{tab:system-system}); Guarded's non-inferiority to Raw on
ATE/RPE; and the accompanying log-level bootstrap 95\% CI. The earlier
30-log bounded consumer remains a historical diagnostic. All other results,
including basin-consistency and the selective-risk scans, are secondary or
exploratory.

\textbf{Chronology.} The system-audit roles and the primary system contract were fixed
before execution. The initial Stage~1 correction tables and the
system audit were run first; the 60-log basin, declared composite-score, and
selective-risk analyses were designed after early results and are therefore
secondary exploratory/post-hoc diagnostics. Later pre-specified observability and
resolution-matched candidate audits were also negative, so they selected no guard threshold or
final rerun. This ordering is a dependency chain:
observability audit $\rightarrow$ candidate generator $\rightarrow$
selective-risk control $\rightarrow$ online system audit.

Every recursive result uses four arms: raw $\hat T$, oracle $T$ (a
non-deployable upper bound), always-correct, and guarded. A single-arm
result is never used to support a recursive claim. Frames are correlated
within a log, so intervals use log-level cluster resampling rather than
per-frame bootstrap, and every calibration/evaluation split is log-disjoint.
Confidence AUROC is computed within each severity tier: pooling tiers would
allow confidence to appear predictive merely by tracking severity instead
of whether a particular correction is trustworthy. The pre-specified gate
requires a 95\% lower bound above chance and a target AUROC of $0.70$; only
then may the confidence score gate Stage~2.

The ablation ladder is correct/mild/severe extrinsic error. Because Stage~1
is undefined without LiDAR, a literal no-LiDAR arm is not meaningful; instead
an edge-vs-random-point control uses the same point budget without
range-discontinuity selection, testing whether the specific geometric signal
drives the correction. If any criterion fails, the result is retained as a
negative result against its pre-specified threshold. We do not relax a
threshold or re-run the final unit to obtain a pass. All runs use fixed
seeds, versioned manifests, and resumable checkpoints; intervals below use
2,000 log-level bootstrap resamples.

The resulting decision sequence is deliberately asymmetric: a candidate may
be rejected by any early audit, but it cannot be admitted by a later positive
metric if an earlier sanity check has failed. This prevents downstream
trajectory numbers from masking a correction that is already unreliable at
the transform level.

\begin{algorithm}[t]
\caption{Audit-gated correction decision}
\label{alg:audit-gate}
\begin{algorithmic}[1]
\REQUIRE Log-disjoint calibration set $D_c$, evaluation set $D_e$;
         candidate corrector $f$; fixed risk targets
\STATE Fit no final threshold on $D_e$; retain the frozen search box and seeds
\STATE Run $f$ from the true extrinsic with zero injected drift on $D_c$
\IF{the zero-injection sanity check shows systematic movement from truth}
\STATE Flag the corrector for a wrong-basin failure-pattern diagnostic; continue only as a diagnostic
\ENDIF
\STATE Measure tier-wise correction improvement and confidence AUROC on $D_c$
\IF{any tier fails the pre-specified correction or AUROC gate}
  \STATE Do not promote $f$ directly to the recursive consumer
\ENDIF
\STATE Select a threshold on $D_c$ under minimum coverage and false-admission risk
\STATE Evaluate raw, oracle, always-correct, and guarded arms on $D_e$
\IF{guarded fails downstream benefit or safety}
  \STATE Run the pre-specified complete system audit as a negative control
  \STATE Freeze the negative result; make no deployment claim
\ELSE
  \STATE Re-test the complete frozen protocol on a fresh log-disjoint unit
\ENDIF
\end{algorithmic}
\end{algorithm}

\section{Status: A First Stage 1 Run (Negative Result)}
\label{sec:status}

We ran Stage~1's estimator as a classical, non-learned optimizer:
candidate LiDAR points at range discontinuities within each scan ring
(a single-frame instantiation of \cite{levinson2013automatic}) are
projected under a candidate extrinsic and scored against an image
edge-distance transform; a bounded search (Powell's method) over a small
$\Delta R,\Delta t$ around the perturbed extrinsic minimizes this score.
The fixed implementation uses a $2$--$55$\,m range window, $>0.5$\,m jump,
2,500-point cap, Canny 50/150 with a $5\times5$ distance transform, Powell
\texttt{maxiter}=60 over $\pm0.24$\,rad/$\pm0.45$\,m, $K=20$, and seed 1337.
Confidence combines the optimizer's relative cost improvement and the
number of edge points used. Every frame is compensated for the
$\sim$10\,ms LiDAR-sweep/camera-shutter time gap via interpolated ego
pose before scoring, so residual ego motion is not mistaken for
miscalibration. Only rotation-and-translation perturbation/correction
were exercised in this run; time-synchronization perturbation was not, as
a static $\Delta T$ cannot repair a frame-pairing error.

\begin{figure*}[t]
\centering
\includegraphics[width=0.92\textwidth]{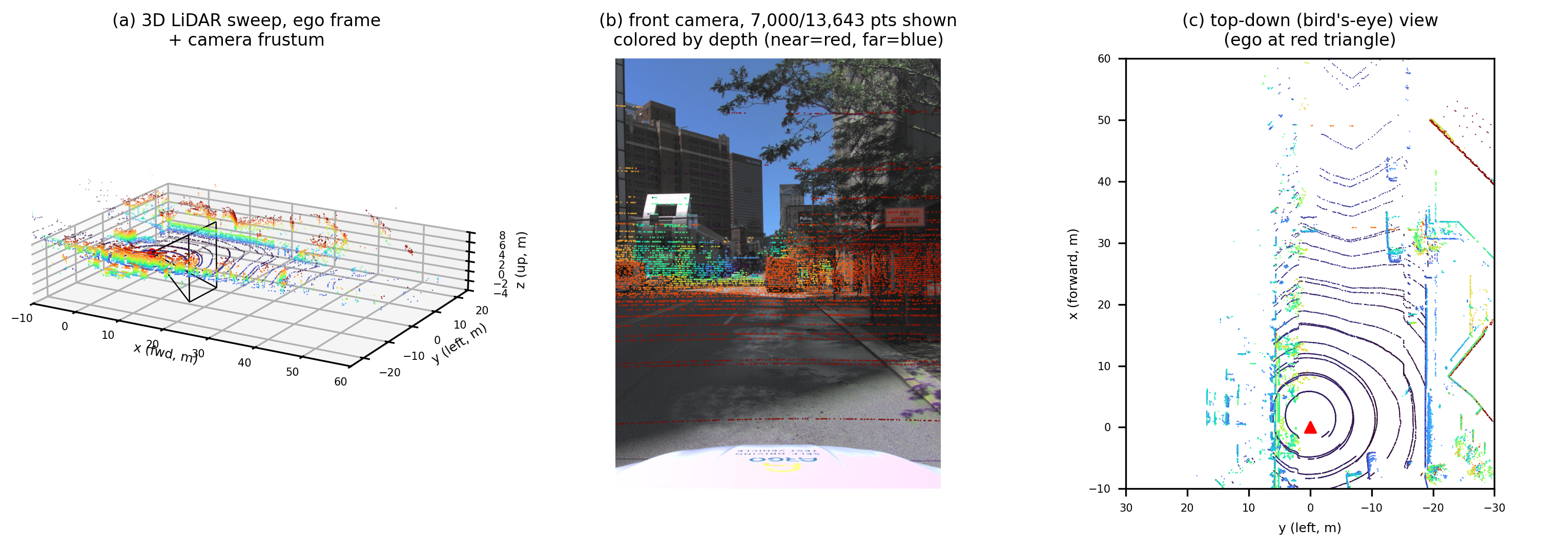}
\caption{One representative Argoverse~2 frame, true extrinsic, no
synthetic drift. (a) LiDAR sweep, height-colored, camera frustum from
calibration. (b) Same sweep projected into the camera image,
depth-colored. (c) Top-down view standing in for a map panel (no HD-map
archive for this log).}
\label{fig:av2_3d}
\end{figure*}

\textbf{Data.} All 27 Argoverse~2 test-split logs available at the time
of this run, the front-center camera only, every third LiDAR frame
(1,435 frames/tier/condition, 5,740 rows total).
Intervals below are log-level cluster bootstraps (2,000 resamples,
resampling whole logs with replacement, not individual frames) rather
than the descriptive 2-log comparison of an earlier pilot; this is the
protocol's formal admission-decision statistic (\S\ref{sec:eval}),
illustrated in Fig.~\ref{fig:av2_3d}.

\begin{table*}[t]
\centering
\scriptsize
\begin{tabular}{@{}llrrrrl@{}}
\toprule
Tier & Points & Raw$\to$Corr. (rot.) & Rot. impr. [95\% CI] & Trans. impr. (m) [95\% CI] & Joint impr. [$s_t{=}1$ m] & AUROC [95\% CI] \\
\midrule
Mild & Edge & $1.25^\circ\!\to\!6.24^\circ$ & $-4.99$ [$-5.47,-4.46$]$^{\dagger}$ & $-0.248$ [$-0.257,-0.239$]$^{\dagger}$ & $-0.268$ [$-0.277,-0.259$]$^{\dagger}$ & $0.512$ [$0.428,0.608$] \\
Mild & Random & $1.25^\circ\!\to\!8.14^\circ$ & $-6.89$ [$-7.58,-6.21$]$^{\dagger}$ & $-0.274$ [$-0.300,-0.250$]$^{\dagger}$ & $-0.306$ [$-0.332,-0.281$]$^{\dagger}$ & $0.298$ [$0.040,0.550$] \\
Severe & Edge & $7.52^\circ\!\to\!7.34^\circ$ & $+0.18$ [$-0.40,0.80$] & $-0.126$ [$-0.136,-0.117$]$^{\dagger}$ & $-0.124$ [$-0.134,-0.115$]$^{\dagger}$ & $0.649$ [$0.605,0.694$]$^{\ddagger}$ \\
Severe & Random & $7.52^\circ\!\to\!9.09^\circ$ & $-1.57$ [$-2.28,-0.86$]$^{\dagger}$ & $-0.165$ [$-0.187,-0.143$]$^{\dagger}$ & $-0.170$ [$-0.194,-0.146$]$^{\dagger}$ & $0.531$ [$0.470,0.594$] \\
\bottomrule
\end{tabular}
\caption{Log-cluster-bootstrap rotation- and translation-error improvement
(raw minus corrected; positive is better), the declared joint error-magnitude
improvement, and within-tier confidence AUROC, 27 logs, 1,435
frames/tier/condition. The joint column uses radians and $s_t=1$ m; it is not
an invariant $SE(3)$ distance. $\dagger$: CI excludes zero in the
\emph{harmful} direction. $\ddagger$: CI lower bound exceeds $0.5$
(significantly better than chance) but not the $0.70$ target. Pooled
(not within-tier) AUROC would read $0.753$ (edge) -- appearing to pass
$0.70$; see discussion below. Translation is significantly harmful in
\emph{every} cell, including severe/edge, where rotation is not
significantly harmed: the simultaneous rotation/translation/joint read is
uniformly negative even where the one-DoF (rotation) read is not.}
\label{tab:stage1results}
\end{table*}

\textbf{None of Stage~1's three criteria are met.} Correction does not
reduce error at both tiers: at \emph{mild} severity it is significantly
harmful for both edge and random points, while at \emph{severe} severity the
point estimate is small and its interval includes zero. Within-tier AUROC
is significantly above chance only for edge points at the severe tier
($0.649$, lower bound $0.605$), still below the $0.70$ target. Edge-selected
points are less harmful than random points at both tiers (severe:
$+0.18^\circ$ versus $-1.57^\circ$), but this does not amount to a
significant net improvement. Translation is significantly harmful in all
four cells, including severe/edge, where rotation is not: the simultaneous
rotation/translation component reading of Table~\ref{tab:stage1results} is
uniformly negative even where the rotation-only reading is ambiguous. This is a negative result for the
single-frame Canny instantiation and does not open Stage~2.

\textbf{The pooled AUROC is a methodological trap.} Pooling both tiers gives
an apparently passing AUROC of $0.753$ for edge points because confidence
tracks severity, not per-instance quality. The within-tier statistic is the
one relevant to a gate, and it does not pass. The fact that this artifact
persists after increasing the data by roughly $13\times$ over an earlier
pilot shows that it is a property of the evaluation design rather than
small-sample noise.

\textbf{Diagnosis.} A direct cost sweep around the true calibration, one
rotation axis at a time, is not reliably unimodal: on two of three axes the
minimum in a $\pm3^\circ$ window is offset $0.5^\circ$--$1^\circ$ from zero,
and on the third the cost decreases toward the window edge. On a diagnostic
27-log subsample, the cost surface favored the true pose over the starting
pose in 77.1\% of mild cases, yet in 93.1\% the optimizer found an even
lower-cost wrong basin and moved $4.60^\circ$ on average for a $1.29^\circ$
injected error. Single outdoor frames contain many Canny edges unrelated to
depth discontinuities (foliage, road texture, and shadows), consistent with
why multi-frame accumulation and learned features are natural remediations.
The result is negative for this classical single-frame cost surface, not
evidence against the staged audit protocol or a learned Stage~1 estimator.

\section{Remediation Attempts}
\label{sec:remediation}

We compare the original single-frame Canny solver with two remediations.
First, multi-frame accumulation jointly optimizes one correction over
$K=20$ frames per log, using one session-level perturbation and a matched
single-frame baseline. Second, a five-convolution fully convolutional network
predicts a depth-discontinuity heatmap from the image using ground-truth-
assisted weak labels on an 18/9 train/holdout partition of the same 27
formal logs. The learned heatmap replaces Canny edges but leaves the optimizer,
search box, and confidence evaluation unchanged. Specifically, the
EdgeLikelihoodNet has channels $3\!\to\!16\!\to\!32\!\to\!32\!\to\!16\!\to\!1$
with dilations $1,2,4,1$, BatchNorm/ReLU hidden blocks, a sigmoid output,
and no downsampling; it is trained with MSE and Adam (learning rate
$10^{-3}$) for 40 epochs, batch size 8, no augmentation, seed 1337, and a
fixed final-epoch checkpoint. Table~\ref{tab:remediation} shows the matched
rotation results.

\begin{table*}[t]
\centering
\scriptsize
\begin{tabular}{@{}llrrrl@{}}
\toprule
Method & Tier & Rot. impr. [95\% CI] & Trans. impr. (m) [95\% CI] & Joint impr. [$s_t{=}1$ m] & AUROC [95\% CI] \\
\midrule
Canny, single-frame$^{a}$ & Mild & $-5.06$ [$-5.80,-4.31$] & $-0.246$ [$-0.260,-0.231$]$^{\ddagger}$ & $-0.268$ [$-0.284,-0.251$] & $0.500$ \\
Canny, single-frame$^{a}$ & Severe & $+0.08$ [$-0.68,0.88$] & $-0.129$ [$-0.144,-0.114$]$^{\ddagger}$ & $-0.128$ [$-0.142,-0.112$] & $0.658$ \\
Canny, multi-frame$^{b}$ & Mild & $-2.62$ [$-3.94,-1.47$] & $-0.331$ [$-0.387,-0.279$]$^{\ddagger}$ & $-0.340$ [$-0.394,-0.294$] & $0.437$ \\
Canny, multi-frame$^{b}$ & Severe & $\mathbf{+2.52}$ [$0.90,4.06$] & $-0.177$ [$-0.244,-0.115$]$^{\ddagger}$ & $-0.158$ [$-0.223,-0.098$] & $\mathbf{0.879}^{\dagger}$ \\
Learned, single-frame$^{a}$ & Mild & $-3.47$ [$-4.15,-2.82$] & $-0.332$ [$-0.354,-0.307$]$^{\ddagger}$ & $-0.341$ [$-0.363,-0.317$] & $0.614$ [$0.517,0.696$] \\
Learned, single-frame$^{a}$ & Severe & $\mathbf{+2.21}$ [$1.44,2.84$] & $-0.193$ [$-0.211,-0.174$]$^{\ddagger}$ & $-0.176$ [$-0.195,-0.157$] & $0.641$ [$0.532,0.743$] \\
Learned, multi-frame$^{c}$ & Mild & $-0.26$ [$-0.83,0.30$] & $-0.329$ [$-0.395,-0.252$]$^{\ddagger}$ & $-0.327$ [$-0.391,-0.250$] & $0.600$ [$0.000,1.000$] \\
Learned, multi-frame$^{c}$ & Severe & $\mathbf{+5.22}$ [$4.62,5.95$] & $-0.139$ [$-0.302,0.004$] & $-0.113$ [$-0.272,+0.027$] & n/a \\
\bottomrule
\end{tabular}
\caption{Rotation-, translation-, and declared joint error-magnitude improvement,
and confidence AUROC,
bold where the 95\% CI excludes zero in the helpful direction (rotation
improvement) or the lower bound exceeds $0.5$ (AUROC). $\dagger$: the only
cell clearing the pre-specified $0.70$ AUROC target. $\ddagger$: translation
CI excludes zero in the \emph{harmful} direction -- true for every cell
except learned multi-frame/severe, whose interval is the widest ($n{=}9$
logs) and merely touches zero. Both cells with a bold (significant) rotation
  gain -- Canny and learned multi-frame/severe -- have significantly worse
translation at the same time: the component-wise read finds no cell in this table
that is unambiguously helpful. The declared joint endpoint is included in
the table; its definition is given in Section~\ref{sec:method}, and the
pre-specified scale sensitivity is summarized below.
$a$: per-frame perturbation, 473
frames on the same 9 log-disjoint logs used to evaluate the learned model
(Canny row recomputed on this subset for a matched comparison; the
27-log Canny numbers in Table~\ref{tab:stage1results} are similar but not
identical). $b$: per-log perturbation, 27 logs (different unit; see
text), single-frame row uses the same per-log perturbation and first
frame only, isolating the effect of accumulation. $c$: learned heatmap
with the same $K{=}20$ multi-frame protocol on the 9 held-out logs;
AUROC is undefined for severe because all nine corrections improved, and
the mild interval is exploratory rather than a formal 27-log gate.}
\label{tab:remediation}
\end{table*}

Multi-frame accumulation yields a significant severe-tier correction
($+2.52^\circ$, CI $[0.90,4.06]$) and AUROC $0.879$ (CI
$[0.714,0.986]$), the only cell clearing both pre-specified rotation and
confidence targets. The learned solver also improves the severe tier
($+2.21^\circ$ single-frame; $+5.22^\circ$ multi-frame on 9 held-out logs),
but its confidence remains below the $0.70$ target and the mild tier is
unresolved. Critically, the component and declared-joint analysis shows that
\emph{both} rotation-improving cells make translation worse at the same time
($-0.177$\,m and $-0.139$\,m respectively); their joint improvements are
$-0.158$ and $-0.113$, with the latter interval crossing zero. Thus the
corrector that "passes" on rotation and confidence would still inject a systematically
larger translation error if it were admitted, which is exactly the
failure mode a rotation-only gate cannot see. Neither remediation passes
both tiers on rotation alone, and none passes on the declared composite
criterion,
so Stage~2 remains closed.

On the same 9-log holdout, the learned solver's zero-injection run wandered
more than $0.5^\circ$ on $99.15\%$ of frames and its basin-consistency AUROC
was $0.623$/$0.662$ on mild/severe, below target. This small diagnostic is
consistent with a wrong-basin failure pattern beyond Canny, but does not
justify a gating decision.

As a pre-specified follow-up, the same learned producer was retrained with the
native $\to388\times512\to256\times340$ path. On the 9-log holdout, mild and
severe rotation improvements were $-3.471^\circ$ (AUROC $0.510$) and
$+2.548^\circ$ (AUROC $0.681$), respectively. Although 45 helpful severe
candidates appeared in 8/15 calibration logs, all 916 correct-tier
zero-injection candidates were harmful; this follow-up therefore closed without a guard
or final-log rerun.

The preceding observability audit reused the fixed 15-log calibration
role: all 2,748 observations had complete descriptors and 90 offline-helpful
candidates occurred across 12 logs, but all 916 correct-tier zero-injection
candidates were harmful. Its safety gate therefore failed; no descriptor was
promoted and no downstream rerun was authorized.

\subsection{External-Dataset Geometry Plumbing Diagnostic (nuScenes)}
\label{sec:nuscenes}

We re-implemented the projection chain on nuScenes
\cite{caesar2020nuscenes} to test sensor-convention portability, not to waive
the Stage~1 gates. Unlike Argoverse~2, nuScenes stores orientation as
$[w,x,y,z]$ quaternions and composes sensor-to-ego and ego-to-global
transforms independently. The implementation therefore checks both the
quaternion convention and the sweep accumulation order before any optimizer
is run.

On a fixed nuScenes Mini sample, single-frame projection placed 2,668 of 34,688
input LiDAR points inside the image under the true extrinsic (7.7\% coverage)
and 2,672 under a fixed camera-local perturbation of $1^\circ$ about the optical
axis and 5\,cm along the image-right axis, with a $7.36$-pixel median shift over
the points visible under both transforms.
Accumulating four preceding sweeps through their logged ego poses into the
same reference camera frame under the true extrinsic raised the point count to
173,376 (12,211 in image, 7.0\% coverage).
The lower accumulated coverage reflects the wider spatial footprint of five
sweeps, not a projection failure. These results are a geometric plumbing
check only: no correspondence-confidence or correction model is fit here,
and passing this check does not relax any Stage~1 admission criterion.
\textbf{Cross-domain diagnostic.} We also ran the classical single- and
multi-frame Canny correctors, unchanged from Section~\ref{sec:remediation},
as an external diagnostic baseline on a 170-scene nuScenes calibration role drawn
from an 850-scene Trainval manifest. The correctors were harmful at every
tested threshold; for example, severe multi-frame changed
$7.60^\circ\to9.28^\circ$ (CI $[-2.63,-0.76]$). Starting from the true pose
with zero injected error still lowered the cost while wandering $8$--$10^\circ$
on 92--94\% of scenes. An independent 20-log Argoverse~2 training sample
reproduced the failure: zero of 2,102 frames improved, with degradation CIs
$[-7.42,-6.27]^\circ$ for edge points and $[-9.79,-8.22]^\circ$ for the
random control. These checks support a cross-dataset wrong-basin failure-pattern
diagnostic, not a deployment gate; the original 27 test logs are analyzed
directly below. The remaining nuScenes scenes are not used for threshold
tuning or as a second positive claim.
Because no locked scene-manifest hash exists for this role,
this is not a cross-domain generalization benchmark.

\subsection{Solver-Disagreement Diagnostic: Basin Consistency}
\label{sec:basin-consistency}

Every true-pose comparison so far (\S\ref{sec:status}, above) detects the
failure by comparing the optimizer's solution against the \emph{true}
extrinsic -- information a deployed system never has. We test whether the
failure instead leaves a trace in something a deployed system does have:
disagreement between independent re-solves of the same correction
problem. If the edge-alignment cost surface has one dominant basin,
re-solving from different starting points should converge to (nearly) the
same pose, whether or not that pose is correct. If it has two or more
comparably-deep basins -- the shape the mild-tier diagnosis above
describes -- restarts should disagree more often on exactly the frames
where the canonical solve lands in the wrong one.

For each (log, tier, method) cell, we re-solve the identical correction
problem -- same LiDAR points, same image-edge cost surface, same
classical Canny corrector used throughout this paper -- five additional
times from starting points drawn uniformly across the full search box,
alongside the canonical zero-init solve. We summarize the six resulting
poses by their mean pairwise geodesic rotation angle (\emph{basin
spread}); low spread is the candidate confidence signal, in place of the
cost-drop heuristic used elsewhere in this paper. This extends the
zero-injection check above from 20 to all 60 downloaded Argoverse~2
train/val exploratory-pool logs and adds the mild and severe tiers, reusing the
same disjointness argument: since these are from the official train/val splits and
Table~\ref{tab:remediation}'s 27 logs are test-split, the two samples
cannot share a log ID regardless of which 27 test-split logs were
originally selected.

\begin{table*}[t]
\centering
\small
\begin{tabular}{@{}llrrl@{}}
\toprule
Tier & Method & \% Improved & Basin-consistency AUROC [95\% CI] & Legacy (cost-drop) AUROC [95\% CI] \\
\midrule
Correct & Multi-frame & $0.0\%$ & n/a (no frame improved) & n/a \\
Correct & Single-frame & $0.0\%$ & n/a (no frame improved) & n/a \\
Mild & Multi-frame & $31.7\%$ & $\mathbf{0.673}$ [$0.519,0.811$]$^{*}$ & $0.548$ [$0.403,0.693$] \\
Mild & Single-frame & $5.0\%$ & $0.596$ [$0.458,0.719$] & $0.632$ [$0.203,0.898$] \\
Severe & Multi-frame & $70.0\%$ & $\mathbf{0.770}^{\dagger}$ [$0.620,0.888$]$^{*}$ & $0.603$ [$0.437,0.758$] \\
Severe & Single-frame & $38.3\%$ & $0.623$ [$0.464,0.779$] & $0.563$ [$0.408,0.702$] \\
\bottomrule
\end{tabular}
\caption{Basin-consistency versus the existing cost-drop confidence
heuristic for predicting whether a correction reduced rotation error on all
60 Argoverse~2 train/validation exploratory-pool logs (360 rows; six
restart-solves per row). Bold marks a point estimate above the legacy
heuristic. $*$ denotes a 95\% CI excluding $0.5$; $\dagger$ denotes the
pre-specified $0.70$ target.}
\label{tab:basinconsistency}
\end{table*}

Basin-consistency is above chance on the exploratory pool for mild/multi-frame
($0.673$, CI $[0.519,0.811]$) and severe/multi-frame ($0.770$, CI
$[0.620,0.888]$), while the legacy score gives $0.548$ and $0.603$.
It outperforms the legacy score in every defined row except mild/single-frame,
and is the only signal significantly above chance in this table. The severe
multi-frame result clears the $0.70$ target that the legacy score misses on
this disjoint sample, whereas the formal 27-log unit below reverses that
ranking. This split-dependent reversal is itself evidence that one sample's
admission result is insufficient to establish a general capability. The mild
multi-frame signal is also the strongest mild-tier ranking found in this
study, but it does not make mild correction admissible. Zero injection
improves no log in either method. These results are exploratory and do not
reopen Stage~2; the formal test below checks whether the signal survives on
the original unit.

\subsection{Resolving Contamination on the Formal 27-Log Unit}
\label{sec:formal27}

We re-run the identical zero-injection and basin-consistency protocol on the
27 test-split logs behind Table~\ref{tab:remediation}, reproducing its
severe/multi-frame result within $0.005^\circ$. Starting from the true pose
with no injected drift, $0$ of 27 logs improve under either solver. Thus the
one cell that passed the original admission criteria is consistent with a
wrong-basin failure pattern on its own formal unit, although this does not prove that
every severe-tier improvement is spurious.

\begin{table*}[t]
\centering
\small
\begin{tabular}{@{}llrrl@{}}
\toprule
Tier & Method & \% Improved & Basin-consistency AUROC [95\% CI] & Legacy (cost-drop) AUROC [95\% CI] \\
\midrule
Correct & Multi-frame & $0.0\%$ & n/a (no frame improved) & n/a \\
Correct & Single-frame & $0.0\%$ & n/a (no frame improved) & n/a \\
Mild & Multi-frame & $25.9\%$ & $\mathbf{0.814}^{\dagger}$ [$0.611,0.968$]$^{*}$ & $0.393$ [$0.160,0.632$] \\
Mild & Single-frame & $3.7\%$ & $0.500$ [$0.308,0.692$] & $0.231$ [$0.077,0.417$] \\
Severe & Multi-frame & $74.1\%$ & $0.757^{\dagger}$ [$0.529,0.935$]$^{*}$ & $\mathbf{0.879}^{\dagger}$ [$0.714,0.986$]$^{*}$ \\
Severe & Single-frame & $59.3\%$ & $0.722^{\dagger}$ [$0.489,0.909$] & $0.659$ [$0.437,0.864$] \\
\bottomrule
\end{tabular}
\caption{Basin-consistency vs.\ the existing cost-drop confidence
heuristic on the formal 27-log Argoverse~2 test-split unit (162 rows; 6
restart-solves per row) -- the same logs and perturbations behind
Table~\ref{tab:remediation}. Bold where basin-consistency's point
estimate exceeds the legacy heuristic's. $*$: 95\% CI excludes $0.5$
(significantly above chance). $\dagger$: point estimate exceeds the
pre-specified $0.70$ target (severe/single-frame's CI still crosses
$0.5$, so it does not count as a significant pass despite clearing the
point-estimate target).}
\label{tab:formal27}
\end{table*}

\textbf{Basin-consistency.} On the formal unit, basin-consistency clears
$0.70$ significantly for mild/multi-frame ($0.814$, CI $[0.611,0.968]$)
and severe/multi-frame ($0.757$, CI $[0.529,0.935]$). The mild correction
remains net harmful ($25.9\%$ of logs improved), but the confidence ranking
separates the minority of helpful cases without access to the true pose.
On severe/multi-frame, the legacy score is higher ($0.879$), illustrating
that a single sample cannot establish a general admission rule.

\textbf{Shared-attractor audit.} For each log, a Monte Carlo null holds the
two observed errors fixed while randomizing independent rotation axes. On
the formal unit, $7/27$ logs fall in the $5\%$ tail, Fisher's test gives
$p<10^{-6}$, and the paired rotation difference is smaller than chance on
$20/27$ logs ($p=0.007$). The full spatial audit gives observed distance
$4.990$ versus null $6.314$ (difference $-1.324$, 95\% CI
$[-2.431,-0.286]$). On the disjoint exploratory train/val pool the analogous
values are $5.734$ versus $7.540$ (difference $-1.806$, 95\% CI
$[-3.032,-0.741]$; $22/60$ per-log $p<0.05$). With the declared 1-m scale,
these remain observational, not causal. The pre-specified audit retains a
negative difference at $s_t=0.1,1,10$\,m in both roles (formal
$-2.421,-1.324,-1.294$; exploratory $-2.095,-1.806,-1.796$), but its
magnitude is scale-dependent; these values are reported as a sensitivity
summary rather than as a causal mechanism test.

Lower basin spread also predicts a smaller correct--severe angle on the
60-log pool (Spearman $\rho=0.62$ correct tier and $0.66$ severe tier,
$p<10^{-4}$), a replicated but observational association rather than causal
evidence.

\subsection{Selective Risk and Coverage}
\label{sec:selective-risk}

\textbf{Selective risk.} A post-hoc hash split of the 60-log pool (30
calibration/30 evaluation) scanned both scores under minimum coverage,
$5\%$ false-admission, and log-bootstrap risk constraints. Neither score
had a validated nontrivial threshold; both selected never-admit and accepted
no evaluation window. Both scores have $0/90$ admitted rows, $0/30$ admitted
logs, and $[0,0]$ coverage intervals in both roles; all admitted-row risks are
undefined. Thus AUROC is a ranking statistic, not a selective safety
certificate.

\section{Bounded Visual--LiDAR Odometry Diagnostic}
\label{sec:odometry}

To evaluate recursive-system safety, we ran the pre-specified extension
rather than treating a calibration score as a downstream odometry result. The
consumer is a bounded, project-native AV2-LIMO-lite visual--LiDAR front end:
projected LiDAR points
are associated with LK features and adjacent camera motion is estimated by
RANSAC PnP \cite{graeter2018limo}. This is a bounded visual--LiDAR diagnostic
contract, implemented without ROS, IMU, semantic labels, or hidden ground
truth at decision time. Its output includes relative pose arrays, fixed
1/5/10/25-frame RPE, tracking loss, reprojection RMSE, inlier association,
and online-motion success.

\textbf{Protocol.} The immutable manifest contains 3 development, 15
calibration, and 15 final logs, all log-disjoint, using the front-center
camera and fixed seed 1337. Each bundle requests 100 sweeps, requires at
least 30 usable frames, and targets 50 consecutive frames; one final log is
therefore marked short and the other 14 qualify for the long-horizon target.
Raw bundles are prepared without recorded-pose compensation. In the online
condition, adjacent visual--LiDAR 3D--2D associations feed RANSAC PnP and a
fractional $SE(3)$ transform is applied according to the timestamp gap.
Recorded ego pose is not used by the online decision or compensation path.

Each tier uses four paired arms: Raw, Oracle, Always-correct, and Guarded.
Oracle is a non-deployable diagnostic; Always-correct exposes recursive
pollution; Guarded is the evaluated admission policy, not a validated
deployment mechanism. The calibration guard
uses only observable solver fields. Its safety condition is the one-sided
95\% upper bound on harmful admissions among admitted zero-injection
candidates, while minimum coverage, distinct-log coverage, an injected-tier
admission, and a helpful admission prevent a reject-all policy from passing.

\textbf{Guard lock.} Calibration produced 2,748 candidate observations over
15 logs. No threshold satisfied both the conditional zero-injection safety
constraint and the non-degeneracy requirements, so the locked policy was
\emph{never-admit negative control} ($0/2{,}748$ candidate admissions,
$0/15$ logs, zero-injection upper bound 1.00 because no zero-injection
candidate was admitted). The final matrix below was nevertheless run with
this locked negative control to expose the complete recursive-system risk;
neither its threshold nor any final metric was used to change calibration.

\begin{table}[t]
\centering
\scriptsize
\begin{tabular}{@{}lrrrr@{}}
\toprule
Metric & Raw & Oracle & Always-correct & Guarded \\
\midrule
ATE RMSE (m) & $7.458$ & $4.520$ & $20.474$ & $7.458$ \\
Rotation RMSE (deg) & $6.193$ & $5.975$ & $11.274$ & $6.193$ \\
10-frame RPE rot. (deg) & $3.153$ & $2.854$ & $5.500$ & $3.153$ \\
10-frame RPE trans. (m) & $3.214$ & $2.239$ & $7.245$ & $3.214$ \\
10-frame joint RPE & $3.215$ & $2.240$ & $7.246$ & $3.215$ \\
Reprojection RMSE (px) & $1.973$ & $1.981$ & $1.942$ & $1.973$ \\
Association coverage & $0.165$ & $0.177$ & $0.141$ & $0.165$ \\
Tracking loss (\%) & $0.22$ & $0.23$ & $1.24$ & $0.22$ \\
\bottomrule
\end{tabular}
\caption{Final tier-blind system metrics, averaged within each log
before the 15-log bootstrap. ATE 95\% CIs are Raw
$[5.203,9.855]$, Oracle $[2.404,7.257]$, Always-correct
$[15.108,26.258]$, and Guarded $[5.203,9.855]$ m. The corresponding
10-frame joint-RPE CIs are Raw $[2.228,4.329]$, Oracle
$[1.369,3.391]$, Always-correct $[5.504,9.302]$, and Guarded
$[2.228,4.329]$. Guarded equals Raw because the locked policy admits no
correction.}
\label{tab:system-system}
\end{table}

Table~\ref{tab:system-system} shows that the online consumer and all
requested map-proxy diagnostics are now present, but the result remains
negative: Guarded--Raw ATE is $0.000$ m (95\% CI $[0.000,0.000]$), whereas
Always-correct--Raw is $-13.017$ m ($[-18.639,-7.729]$). The Oracle arm
provides a useful upper reference, but does not rescue the correction
mechanism. Across Raw and Guarded, 3,291 of 3,297 adjacent edges were
estimated online and six used the explicit fallback. No HD-map archive was
available for these bundles, so we report reprojection and association
consistency rather than claiming map consistency.

\begin{table}[t]
\centering
\scriptsize
\begin{tabular}{@{}lrrr@{}}
\toprule
Evaluation item & Calibration & Final & Decision \\
\midrule
Logs / tier rollouts & 15 / 45 & 15 / 45 & log-disjoint \\
Candidate observations & 2,748 & 2,748 & audit only \\
Usable frames & $\geq35$ & $\geq36$ & min. 30 \\
Long-horizon logs ($\geq50$) & -- & 14/15 & 1 short-log \\
Guarded admissions & locked 0 & 0 & negative control \\
\bottomrule
\end{tabular}
\caption{System-audit data and guard results. The final candidate count is shown
only to document the frozen stream; the guard decision was made from the
calibration role, and the final candidate observations were not used for
selection.}
\label{tab:system-guard}
\end{table}

This extension closes the requested measurement gaps but does not support
the stronger system claim: it has complete six-DoF correction/trajectory
metrics, online motion compensation, and a bounded visual--LiDAR diagnostic
consumer,
longer trajectories, and reprojection/association measures, yet no
non-degenerate guarded operating point. We therefore retain the paper's
negative conclusion and do not infer deployment safety, mapping quality, or
closed-loop benefit from the negative-control audit.

\section{Discussion}
\label{sec:discussion}

\textbf{Contribution, given the observed outcome.} We provide a
scope-limited audit protocol that separates correction quality, confidence
ranking, and recursive risk. The single-frame Canny solver fails under mild
drift; multi-frame accumulation recovers a severe-tier rotation point
estimate, but translation worsens simultaneously, and zero-injection and
composite-score audits are consistent with a wrong-basin failure pattern.
The learned solver improves the severe tier on a small holdout, but its
confidence remains below target. Basin-consistency ranks selected cases more
informatively, yet the post-hoc selective-risk audit finds no validated
nonzero-coverage gate. Thus correction reliability and gating reliability are
separate: a useful ranking signal need not support a safe threshold.

The system-level result strengthens this interpretation. Among 2,748
calibration candidates over 15 logs, the conditional zero-injection safety
and non-degeneracy requirements have no common operating point. The locked
negative control therefore admits no correction on the independent final
role, so the present stream supports neither a safe threshold nor a
downstream improvement claim.

\textbf{Recursive relevance.} In the visual--LiDAR odometry consumer, a stale
transform can bias feature association repeatedly. Across 15 final logs and
45 tier rollouts, the locked guard admitted nothing and matched Raw (ATE
7.458\,m; 10-frame joint RPE 3.215), whereas Always-correct was worse
(20.474\,m; 7.246). This supports auditing recursive pollution, not a claim
that the guard improves odometry.

\textbf{Audit implication.} Future studies should gate zero-injection,
tier-wise confidence, and selective risk before the recursive consumer;
here the severe rotation signal fails on translation, basin ranking has no
selective threshold, and Guarded=Raw is not an improvement.

\textbf{Threats to validity and reporting discipline.} We keep distinct
roles rather than pooling evidence: 27 test logs are the formal Stage~1 unit,
60 train/validation logs are exploratory, nine held-out logs compare learned
signals, and the system audit uses separate 15-log calibration and final
roles. The 170-scene nuScenes sample is diagnostic, not frozen transfer.
This separation limits external validity but prevents exploratory results
from becoming admission claims. Holm--Bonferroni hardening and log-cluster
intervals/MDEs (Appendix~\ref{app:hardening}) are reported separately from ranking, selective-risk, and
deployment claims.

\textbf{Limitations and implications.} We use synthetic drift and one AV2
camera. The 30--100-frame consumer has no HD-map archive, so
reprojection/association are map-independent proxies; degenerate PnP edges
can fall back. Time-offset and organic mounting drift remain outside scope,
and the 12-log pose/no-motion audit is exploratory. The 1-m spatial score is
task-scaled, not a dimensionless certificate; AUROC is ranking, not safety.
The $0.1/1/10$-m audit preserves the sign but changes magnitude and tail
counts. A successor must re-open the same log-disjoint gates before any
deployment or closed-loop claim.

\section{Conclusion}
We introduced a confidence-gated camera--LiDAR correction protocol and
separated correction quality, confidence ranking, and recursive risk.
Multi-frame accumulation improves severe-tier rotation but worsens translation
and fails zero-injection; learned correspondence retains a severe-tier gain on
a small holdout, but its confidence misses the admission target. The
resolution-matched audit also fails zero-injection, so no candidate reopens the
recursive branch.

On the formal 27-log unit, basin-consistency clears the $0.70$ ranking target
for multi-frame conditions; the declared composite score is $4.990$ versus
$6.314$ under its independent-direction null. The exploratory pool gives
$5.734$ versus $7.540$ (difference $-1.806$, 95\% CI
$[-3.032,-0.741]$). These are observational, use a declared 1-m translation
scale, and do not identify a causal basin. The pre-specified $0.1/1/10$-m audit
keeps the sign but changes magnitude and tail counts; the nuScenes and
Argoverse~2 zero-injection checks support a failure-pattern diagnostic, not
causal or generalization claims.

A calibration-selected guard has no validated nonzero-coverage operating
point. In the extended consumer, the locked negative control therefore admits
nothing and matches Raw, whereas unconditional correction is harmful. Even
with auxiliary six-DoF, online-motion, long-horizon, reprojection, and
association audits, the recursive branch lacks a safe correction gate. A
future system would need to pass the same tier-wise correction, confidence,
zero-injection, and downstream benefit/safety gates before claiming deployment
or closed-loop benefit. Until then, the evidence
supports auditing correction, confidence, and recursive contamination
separately, but not a deployable calibration gate, downstream odometry
improvement, or closed-loop safety claim.

\appendices
\input{appendix}

\bibliographystyle{IEEEtran}
\bibliography{ref}

\end{document}

%% file: appendix.tex
\section{Data Contract}
\label{app:data-contract}

Table~\ref{app:tab:datacontract} lists every dataset role used in the main
text and this appendix, with its official split, camera, sampling unit,
perturbation rule, log/scene count, and split identity. Two
intersections are easy to misread and are stated explicitly here rather
than only in prose: (i) the learned corrector's 18-log training role and
9-log evaluation role are not a separate sample -- they are a fixed
partition of the same 27 formal test-split logs used throughout Stage~1;
and (ii) the selective-risk audit's 30/30 calibration/evaluation split is
not a new sample either -- it is a post-hoc partition of the same 60-log
exploratory pool used for the exploratory basin-consistency and
full-$SE(3)$ results. Neither partition was chosen after inspecting the
partition's own held-out outcomes.

\begin{table*}[t]
\centering
\scriptsize
\setlength{\tabcolsep}{2pt}
\begin{tabular}{@{}P{0.16\linewidth}P{0.17\linewidth}P{0.24\linewidth}P{0.15\linewidth}P{0.24\linewidth}@{}}
\toprule
Role & Dataset / split & Unit, stride, perturbation & N & Split identity \\
\midrule
Stage~1 formal (Tables~\ref{tab:stage1results} and \ref{tab:remediation} Canny; formal basin-consistency) &
AV2 Sensor Dataset, official \textbf{test} split &
per-frame, every 3rd LiDAR frame; 3 tiers (correct/mild/severe) sampled once per (log, frame, tier) via a deterministic hash seed &
27 logs, 1{,}435 frames/tier/condition &
Pre-specified 27-log test-split list \\
Learned-corrector training role &
Same 27 formal logs (fixed 18-log subset) &
per-frame, same tier sampling as above &
18 of the 27 logs above &
Fixed log-level training partition \\
Learned-corrector evaluation role &
Same 27 formal logs (disjoint 9-log subset) &
per-frame, same tier sampling as above &
9 of the 27 logs above; 473 frames/tier &
Fixed log-level evaluation partition \\
AV2 60-log exploratory pool (basin-consistency, full-$SE(3)$ exploratory) &
AV2 Sensor Dataset, official \textbf{train}+\textbf{val} splits (30 train, 15 val\_calibration, 15 val\_eval) &
per-frame, same tier sampling as above &
60 logs &
Pre-specified train/validation log membership and source-split labels \\
Selective-risk calibration / evaluation split &
Post-hoc 30/30 partition of the same 60-log pool above (no new logs) &
same as the 60-log pool &
30 + 30 &
Post-hoc calibration/evaluation partition of the same pool \\
System audit: development, calibration, final &
AV2 Sensor Dataset, official train and validation splits &
Requested 100 sweeps; at least 30 usable consecutive frames and a 50-frame target; 3 tiers sampled once per log and evaluated with four paired arms &
3 + 15 + 15 = 33 logs &
Fixed role manifest with log-disjoint development, calibration, and final roles \\
Exploratory motion-compensation audit &
AV2 Sensor Dataset, separate from the final system audit &
Recorded-pose, no-motion, and synthetic-noisy-pose conditions on otherwise matched bundles &
12 logs &
Exploratory diagnostic only; not used to select the final guard \\
nuScenes external diagnostics (projection chain; classical Canny correctors) &
NuScenes Mini (projection chain); NuScenes Trainval (correctors) &
Fixed Mini sample; 170-scene calibration role drawn from an 850-scene manifest, with correctors unchanged from the main text &
Mini sample; 170 of 850 scenes &
Diagnostic only; no locked scene-manifest hash \\
nuScenes system-audit transfer &
NuScenes Trainval &
Transfer was deferred because the Argoverse~2 system audit did not pass its locked safety and benefit gates &
Not run &
No cross-domain claim is made \\
\bottomrule
\end{tabular}
\caption{Data contract for every dataset role referenced in the main text
and this appendix. All Argoverse~2 rows use the
\texttt{ring\_front\_center} camera.
The split identity denotes a pre-specified log-level membership record; no
final metric is used to change role assignment.
}
\label{app:tab:datacontract}
\end{table*}

\section{Auxiliary Six-DoF Endpoint}
\label{app:six-dof}

The historical scalar Stage~1 artifacts preserve rotation and translation
error magnitudes but not the corrected pose itself.  For that fixed
error-magnitude reanalysis we therefore report the declared composite, not a
full-pose geodesic or a pixel-space reprojection metric; the independently
rerun pose-persisted audit is reported in Appendix~\ref{app:pixel-audit}.
Rotation values are converted from degrees
to radians before combining with translation at $s_t=1$\,m. All intervals
resample complete logs (2,000 draws); positive improvement means raw error
minus corrected error.

\begin{table*}[t]
\centering
\scriptsize
\begin{tabular}{@{}llrrrr@{}}
\toprule
Source & Tier & Logs & Rotation [95\% CI] & Translation (m) [95\% CI] & Joint ($s_t{=}1$ m) [95\% CI] \\
\midrule
Table~\ref{tab:stage1results} edge & Mild & 27 & $-4.99$ [$-5.47,-4.46$] & $-0.248$ [$-0.257,-0.239$] & $-0.268$ [$-0.277,-0.259$] \\
Table~\ref{tab:stage1results} random & Mild & 27 & $-6.89$ [$-7.58,-6.21$] & $-0.274$ [$-0.300,-0.250$] & $-0.306$ [$-0.332,-0.281$] \\
Table~\ref{tab:stage1results} edge & Severe & 27 & $+0.18$ [$-0.40,+0.80$] & $-0.126$ [$-0.136,-0.117$] & $-0.124$ [$-0.134,-0.115$] \\
Table~\ref{tab:stage1results} random & Severe & 27 & $-1.57$ [$-2.28,-0.86$] & $-0.165$ [$-0.187,-0.143$] & $-0.170$ [$-0.194,-0.146$] \\
\midrule
Canny single & Mild & 9 & $-5.06$ [$-5.80,-4.31$] & $-0.246$ [$-0.260,-0.231$] & $-0.268$ [$-0.284,-0.251$] \\
Canny single & Severe & 9 & $+0.08$ [$-0.68,+0.88$] & $-0.129$ [$-0.144,-0.114$] & $-0.128$ [$-0.142,-0.112$] \\
Canny multi & Mild & 27 & $-2.62$ [$-3.94,-1.47$] & $-0.331$ [$-0.387,-0.279$] & $-0.340$ [$-0.394,-0.294$] \\
Canny multi & Severe & 27 & $+2.52$ [$+0.90,+4.06$] & $-0.177$ [$-0.244,-0.115$] & $-0.158$ [$-0.223,-0.098$] \\
Learned single & Mild & 9 & $-3.47$ [$-4.15,-2.82$] & $-0.332$ [$-0.354,-0.307$] & $-0.341$ [$-0.363,-0.317$] \\
Learned single & Severe & 9 & $+2.21$ [$+1.44,+2.84$] & $-0.193$ [$-0.211,-0.174$] & $-0.176$ [$-0.195,-0.157$] \\
Learned multi & Mild & 9 & $-0.26$ [$-0.83,+0.30$] & $-0.329$ [$-0.395,-0.252$] & $-0.327$ [$-0.391,-0.250$] \\
Learned multi & Severe & 9 & $+5.22$ [$+4.62,+5.95$] & $-0.139$ [$-0.302,+0.004$] & $-0.113$ [$-0.272,+0.027$] \\
\bottomrule
\end{tabular}
\caption{Auxiliary six-DoF error-magnitude reanalysis of the Stage~1
tables. The joint column is dimensionless normalized scale units from
$\sqrt{\theta_{\rm rad}^2+(t/1\,\mathrm{m})^2}$; it is not a claim of an
invariant $SE(3)$ distance. A 12-cell rotation check reproduced the
point estimates of the main tables before the additional columns were computed.}
\label{app:tab:six-dof}
\end{table*}

\section{Statistical Hardening}
\label{app:hardening}

To strengthen the statistical audit without changing any data role or
decision, we reanalyzed the fixed twelve-cell grid at the log level.  Each
frame-level endpoint was first averaged within log; 10,000 complete-log
bootstrap draws provide the intervals below, and two-sided sign-flip tests
are adjusted by Holm--Bonferroni separately within each endpoint family.
Positive values are improvements.  The table reports the mean and 95\% CI,
followed by the adjusted $p$-value in parentheses.

\begin{table*}[t]
\centering
\scriptsize
\setlength{\tabcolsep}{2pt}
\begin{tabular}{@{}lllrllrllr@{}}
\toprule
Source & Tier & Variant & \multicolumn{2}{c}{Rotation (deg)} &
\multicolumn{2}{c}{Translation (m)} & \multicolumn{2}{c}{Joint ($s_t{=}1$ m)} \\
\cmidrule(lr){4-5}\cmidrule(lr){6-7}\cmidrule(l){8-9}
 & & & Mean [95\% CI] & $p_H$ & Mean [95\% CI] & $p_H$ & Mean [95\% CI] & $p_H$ \\
\midrule
Formal Canny & Mild & edge & $-4.974$ [$-5.462,-4.464$] & .006 & $-0.247$ [$-0.256,-0.238$] & .006 & $-0.268$ [$-0.277,-0.258$] & .006 \\
Formal Canny & Mild & control & $-6.885$ [$-7.584,-6.215$] & .006 & $-0.274$ [$-0.303,-0.250$] & .006 & $-0.306$ [$-0.335,-0.281$] & .006 \\
Formal Canny & Severe & edge & $+0.194$ [$-0.387,+0.814$] & 1.000 & $-0.126$ [$-0.136,-0.117$] & .006 & $-0.124$ [$-0.134,-0.115$] & .006 \\
Formal Canny & Severe & control & $-1.569$ [$-2.297,-0.847$] & .006 & $-0.165$ [$-0.190,-0.143$] & .006 & $-0.170$ [$-0.195,-0.147$] & .006 \\
Held-out Canny & Mild & edge & $-5.064$ [$-5.835,-4.325$] & .031 & $-0.246$ [$-0.260,-0.231$] & .012 & $-0.268$ [$-0.284,-0.251$] & .018 \\
Held-out Canny & Severe & edge & $+0.072$ [$-0.696,+0.851$] & 1.000 & $-0.129$ [$-0.144,-0.114$] & .020 & $-0.128$ [$-0.142,-0.111$] & .018 \\
Canny multi & Mild & $K{=}20$ & $-2.625$ [$-3.926,-1.459$] & .006 & $-0.331$ [$-0.384,-0.277$] & .006 & $-0.340$ [$-0.389,-0.292$] & .006 \\
Canny multi & Severe & $K{=}20$ & $+2.522$ [$+0.951,+4.082$] & .040 & $-0.177$ [$-0.245,-0.115$] & .006 & $-0.158$ [$-0.224,-0.097$] & .007 \\
Learned single & Mild & edge & $-3.462$ [$-4.161,-2.811$] & .031 & $-0.332$ [$-0.354,-0.308$] & .020 & $-0.341$ [$-0.363,-0.316$] & .020 \\
Learned single & Severe & edge & $+2.218$ [$+1.480,+2.841$] & .040 & $-0.193$ [$-0.210,-0.174$] & .020 & $-0.176$ [$-0.194,-0.158$] & .020 \\
Learned multi & Mild & $K{=}20$ & $-0.263$ [$-0.842,+0.310$] & 1.000 & $-0.329$ [$-0.396,-0.250$] & .020 & $-0.327$ [$-0.393,-0.247$] & .020 \\
Learned multi & Severe & $K{=}20$ & $+5.219$ [$+4.602,+5.967$] & .028 & $-0.139$ [$-0.301,+0.014$] & .138 & $-0.113$ [$-0.270,+0.034$] & .186 \\
\bottomrule
\end{tabular}
\caption{Log-level statistical hardening. $p_H$ is Holm--Bonferroni
adjusted within the twelve-cell family for the displayed endpoint; the
three endpoint families are not pooled. The prospective 80\%-power MDE
ranges across the grid are $0.728$--$2.271^\circ$ for rotation,
$0.013$--$0.239$\,m for translation, and $0.014$--$0.234$ joint units.
MDE is a precision diagnostic, not post-hoc achieved power.}
\label{app:tab:hardening}
\end{table*}

The severe Canny multi-frame rotation gain survives the declared family
correction, but its simultaneous translation and joint improvements remain
negative.  Thus the stronger statistical accounting reinforces the existing
negative safety conclusion rather than reopening the correction guard.
AUROC is still a tier-stratified ranking gate, not an application-independent
safety guarantee; the $0.70$ value is retained as a screening target.

\section{Pose-Persisted Pixel Audit}
\label{app:pixel-audit}

The producer reruns persist the complete raw and corrected pose rather than
only scalar pose errors.  We use those poses to reconstruct the same
time-compensated point sets and compare projections with the true calibration.
The 3-pixel column is the fraction within 3 pixels of a Canny edge and is an
edge-consistency proxy, not a visual feature-association accuracy measure.
The final column is the median corrected-to-true pixel displacement on common
input-point identities.

\begin{table*}[t]
\centering
\scriptsize
\setlength{\tabcolsep}{3pt}
\begin{tabular}{@{}lllrrrrr@{}}
\toprule
Source & Tier & Variant & Corrected RMSE (px) & True RMSE (px) &
Corrected 3px & True 3px & Corr.-true median (px) \\
\midrule
Formal Canny & Mild & edge & 12.13 & 15.32 & .5323 & .4639 & 147.13 \\
Formal Canny & Mild & control & 21.26 & 35.90 & .4473 & .3277 & 217.25 \\
Formal Canny & Severe & edge & 12.48 & 15.32 & .5257 & .4639 & 167.49 \\
Formal Canny & Severe & control & 21.69 & 35.90 & .4410 & .3277 & 237.72 \\
Canny multi & Mild & $K{=}20$ & 14.32 & 15.21 & .4793 & .4644 & 92.57 \\
Canny multi & Mild & single & 10.58 & 13.33 & .5457 & .4640 & 148.77 \\
Canny multi & Severe & $K{=}20$ & 14.85 & 15.21 & .4818 & .4644 & 107.70 \\
Canny multi & Severe & single & 10.85 & 13.33 & .5336 & .4640 & 153.08 \\
Learned single & Mild & edge & 19.05 & 16.75 & .3896 & .4347 & 113.15 \\
Learned single & Severe & edge & 19.34 & 16.75 & .3890 & .4347 & 117.12 \\
Learned multi & Mild & $K{=}20$ & 17.62 & 16.84 & .4103 & .4348 & 28.95 \\
Learned multi & Mild & single & 13.84 & 13.02 & .4250 & .4614 & 49.19 \\
Learned multi & Severe & $K{=}20$ & 17.55 & 16.84 & .4093 & .4348 & 30.80 \\
Learned multi & Severe & single & 14.17 & 13.02 & .4556 & .4614 & 115.25 \\
\bottomrule
\end{tabular}
\caption{Pose-persisted Stage~1 pixel audit on the same rows as the
correction outputs. RMSE and 3-pixel rates are evaluated against the Canny
distance transform for each image; they are not independent feature-track
association metrics. Large corrected-to-true displacement shows that a
cost-surface solution can improve edge proximity while moving away from the
true calibration.}
\label{app:tab:pixel-audit}
\end{table*}

\section{Independent Association-Label Audit}
\label{app:public-association}

To determine whether the missing association labels could be recovered from
public calibration resources, we audited a checkerboard evaluator fixture, a
target-based correspondence export, and a held-out camera--LiDAR detection
cache.  The first uses the checkerboard configuration and template released
with the Vicon calibration software~\cite{charron2025viconcalibration}; it
passes a 30-keypoint projection smoke and a deterministic TP/FP/FN metric
case.  This is a software check, not a natural-scene benchmark.

The target-based example from Xiaoliang~\cite{xiaoliang2019lidarcameracalibration}
contains 36 image/cloud frames, 48 canonical image corner points, 100 LiDAR
target points per frame, and 33 frames with both camera and LiDAR target
models.  These files establish frame-level target pairing, but they do not
provide shared point identities between the camera and LiDAR observations.

The public xcalib-A9 cache~\cite{xcaliba9} supplies an independent label
format for 2D camera detections and 3D LiDAR detections.  We used its test
cache without fitting thresholds.  For transparency, the all-pairs row in
Table~\ref{app:tab:public-association} predicts every camera--LiDAR pair; it
is a diagnostic lower bound, not a prediction from the proposed corrector.

\begin{table*}[t]
\centering
\scriptsize
\setlength{\tabcolsep}{3pt}
\begin{tabular}{@{}P{0.18\textwidth}P{0.15\textwidth}P{0.32\textwidth}P{0.30\textwidth}@{}}
\toprule
Audit role & Unit & Observed data or metric & Interpretation \\
\midrule
Vicon checkerboard & 30 keypoints & Projection finite; GT ID pairs
TP/FP/FN $=30/0/0$; adversarial case $=29/2/1$ & Evaluator smoke only \\
Xiaoliang target example & 36 image/cloud frames & 48 image points; 100 LiDAR
points/frame; 33 paired target-model frames & Target-frame format; no
point-level cross-sensor IDs \\
A9 north & 29 sensor-frames & 691 camera detections; 410 LiDAR detections;
319 labels; all-pairs precision $0.0321$, recall $1.0$ & Detection-level label
audit \\
A9 south & 29 sensor-frames & 699 camera detections; 525 LiDAR detections;
0 labels; all-pairs precision $0$, recall n/a & No positive labels in this
channel \\
A9 total & 58 sensor-frames & 1,390 camera detections; 935 LiDAR detections;
319 labels; all-pairs precision $0.0141$, recall $1.0$ & Not method P/R and
not feature-track P/R \\
\bottomrule
\end{tabular}
\caption{Independent association-label audit.  The A9 all-pairs values are
computed against its published match matrices and are included to expose the
label accounting scale, not to claim performance for the proposed method.}
\label{app:tab:public-association}
\end{table*}

These checks improve reproducibility of the evaluator and provide a
detection-level association reference, but they do not close the original
natural-scene feature-track limitation.  The Stage~1 Canny quantity therefore
remains an edge-consistency proxy rather than a feature-association
precision/recall measurement.

The ranked follow-up experiments are summarized in
Table~\ref{app:tab:ranked-association}.  The xiaoliang rows use real camera
images and independently detected checkerboard corners; their raw and
corrected poses are controlled perturbations of a per-frame PnP reference.
Consequently, they validate the target-image evaluator but do not constitute
an independent camera--LiDAR feature-track experiment.  The A9 baseline uses
the train/validation-locked threshold and evaluates the test cache once.

\begin{table*}[t]
\centering
\scriptsize
\setlength{\tabcolsep}{3pt}
\begin{tabular}{@{}P{0.22\textwidth}P{0.17\textwidth}P{0.20\textwidth}P{0.25\textwidth}P{0.10\textwidth}@{}}
\toprule
Experiment / role & Unit & Reprojection RMSE (px) & Association P/R/F1 & Coverage \\
\midrule
Synthetic AprilGrid-like / raw & 1 frame, 46 reference pairs & 22.584 & 0/0/0 & 0 \\
Synthetic AprilGrid-like / corrected & 1 frame, 46 reference pairs & 3.068 & 0.9111/0.8913/0.9011 & 0.8913 \\
Synthetic AprilGrid-like / true & 1 frame, 46 reference pairs & 0.504 & 1/1/1 & 1 \\
Xiaoliang checkerboard / raw & 32 real target images, 1,536 pairs & 38.168 & 0/0/0 & 0 \\
Xiaoliang checkerboard / corrected & 32 real target images, 1,536 pairs & 3.897 & 1/1/1 & 1 \\
Xiaoliang checkerboard / true PnP reference & 32 real target images, 1,536 pairs & 0.609 & 1/1/1 & 1 \\
A9 geometric baseline / test micro & 58 sensor-frames, 319 labels & -- & 0.3761/0.9279/0.5353 & 0.9279 \\
A9 geometric baseline / north & 29 sensor-frames, 319 labels & -- & 0.8481/0.9279/0.8862 & 0.9279 \\
A9 geometric baseline / south & 29 sensor-frames, 0 labels & -- & 0/--/-- & --; 438 FP \\
Synthetic tolerance sweep & 216 evaluator cells & -- & Monotonicity checks passed & -- \\
\bottomrule
\end{tabular}
\caption{Results of the ranked association experiments.  Target rows are
controlled evaluator diagnostics; the A9 rows are detection-level baseline
results.  A9 threshold candidates came from train and the locked threshold
$0.80478728$ was selected on validation.  No row is a natural-scene
feature-track result or a Stage~1 method gain.}
\label{app:tab:ranked-association}
\end{table*}

\section{Selective-Risk Endpoint}
\label{app:selective-risk}

The post-hoc selective-risk audit uses the same 60-log exploratory pool as
the basin analysis, with a fixed 30-log calibration role and 30-log held-out
role. Thresholds were selected on calibration only using minimum five
admissions, false-admission rate at most $5\%$, and an upper selective-risk
bound of zero; uncertainty uses 10,000 complete-log bootstrap draws. The
selected threshold for both scores is the explicit never-admit sentinel.

\begin{table*}[t]
\centering
\scriptsize
\begin{tabular}{@{}lccccc@{}}
\toprule
Score & Role & Decision / admitted rows & Admitted logs & Coverage [95\% CI] & Helpful / false / risk \\
\midrule
Basin spread & Calibration & never-admit / 0 of 90 & 0 of 30 & $0\%$ [$0,0$] & n/a / n/a / n/a \\
Basin spread & Held-out & never-admit / 0 of 90 & 0 of 30 & $0\%$ [$0,0$] & n/a / n/a / n/a \\
Legacy cost drop & Calibration & never-admit / 0 of 90 & 0 of 30 & $0\%$ [$0,0$] & n/a / n/a / n/a \\
Legacy cost drop & Held-out & never-admit / 0 of 90 & 0 of 30 & $0\%$ [$0,0$] & n/a / n/a / n/a \\
\bottomrule
\end{tabular}
\caption{Complete selective-risk endpoint. The final column is
helpful-admission rate / false-admission rate / selective rotation risk;
translation and joint risk are also undefined because no row is admitted.
The held-out rows are evaluated once using the calibration-selected decision;
no final outcome is used for threshold selection.}
\label{app:tab:selective-endpoint}
\end{table*}

\section{Qualitative nuScenes Projection Figure}
\label{app:nuscenes-qual}

\begin{figure*}[t]
\centering
\includegraphics[width=0.85\textwidth]{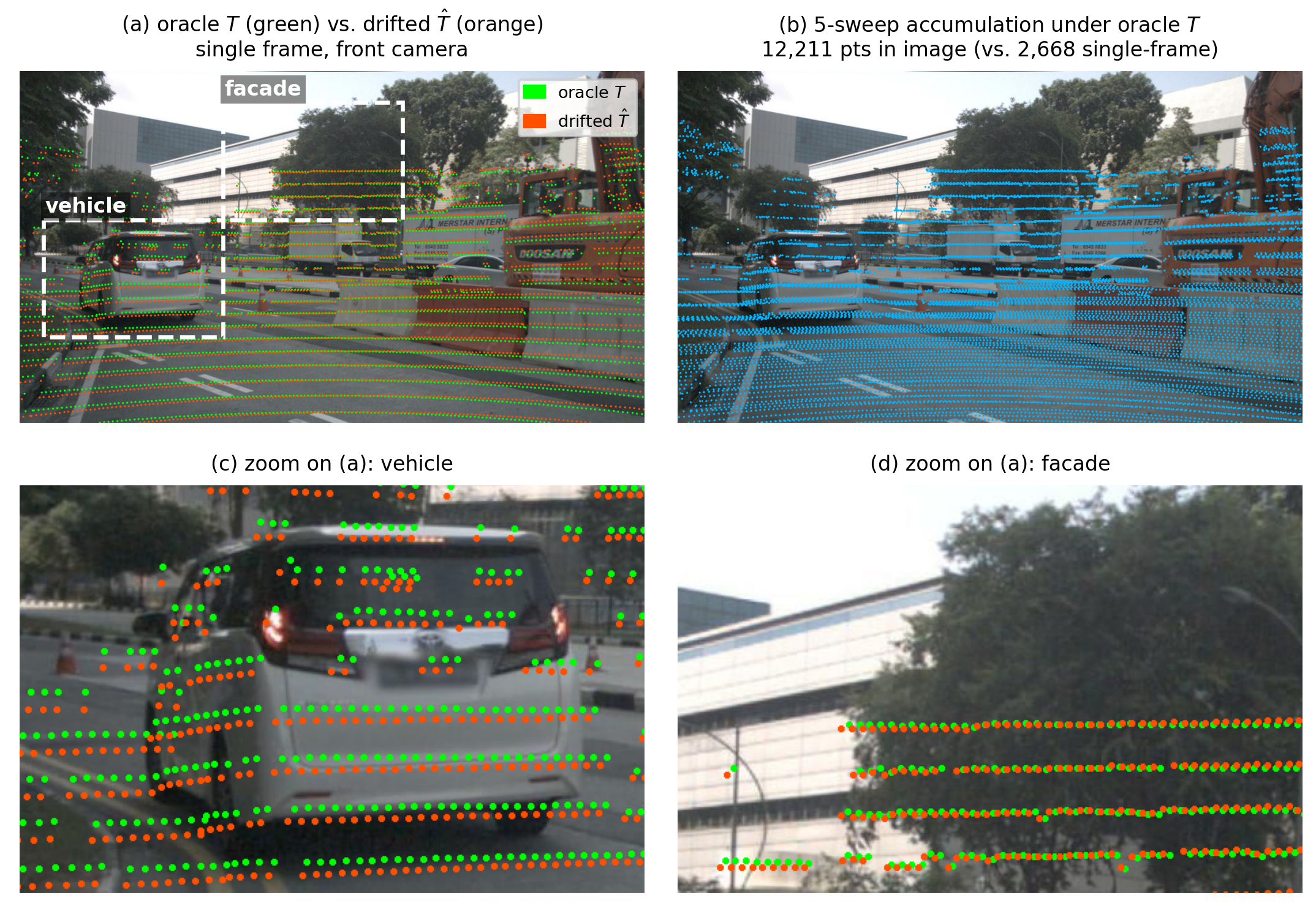}
\caption{Qualitative camera--LiDAR projection on one representative real
nuScenes Mini frame, front camera. (a) LiDAR points projected under the
oracle extrinsic $T$ (green) and under a $1^\circ$/5\,cm synthetic drift
$\hat{T}$ (orange), single frame; dashed boxes mark the two crops shown
below. (b) Points from 5 accumulated LiDAR sweeps projected into the same
camera frame under $T$ (cyan), $4.6\times$ the single-frame point count.
(c)/(d) Zoomed crops of (a) on a vehicle silhouette and a building facade.
The offset between the oracle and drifted point rows is larger in (c), away
from the image center, than in (d), as expected for a rotation about the
optical axis; its median over all points is the $\sim$7-pixel shift reported
in the main text's nuScenes subsection. No correction or confidence model is applied here; this figure
illustrates the projection/coverage check only, not a Stage~1 admission
result.}
\label{app:fig:nuscenes_qual}
\end{figure*}

The main text's nuScenes geometric-portability check reports the finding
numerically (a $7.36$-pixel single-frame median shift for a $1^\circ$/5\,cm
perturbation, and the in-image point count of a 5-sweep accumulation). This is not a Stage~1
correction or admission result -- no correspondence-confidence or correction
model is fit here -- but the same shift is shown directly in
Fig.~\ref{app:fig:nuscenes_qual} for a reader who wants to see it rather than
take the pixel figure on faith.

\clearpage
\onecolumn
\section{Reproducibility Parameters}
\label{app:reproducibility}

Table~\ref{app:tab:repro} lists the numeric parameters that determine every
correction or gate reported in the main text, so that a reader does not
have to reconstruct them from source. All values are read directly from
the frozen implementation.

\begin{longtable}{@{}P{0.28\linewidth}P{0.68\linewidth}@{}}
\caption{Reproducibility parameters for every correction and gate reported
in the main text.}
\label{app:tab:repro}\\
\toprule
Parameter & Value \\
\midrule
\endfirsthead
\toprule
Parameter & Value \\
\midrule
\endhead
\bottomrule
\endfoot
\endlastfoot
Candidate LiDAR points (``edge'') &
Range-discontinuity points: adjacent, within the same scan ring in azimuth
order, to a range jump $>0.5$\,m; range window $2$--$55$\,m; capped at
2{,}500 points, subsampled uniformly at random if exceeded. \\
Random-point control &
Same range window and point budget as the edge set, drawn uniformly at
random with no discontinuity selection. \\
Image cost surface &
Canny edge map (OpenCV, thresholds 50/150) on grayscale, inverted, then an
L2 distance transform with a $5\times5$ mask. \\
Depth / occlusion filtering &
A candidate point is kept only if it is in front of the camera
($z>0.5$\,m) and within a normalized radius $<1.2$ (bounds AV2's radial
distortion polynomial to its well-behaved range). If fewer than 20 points
survive filtering, the cost function returns a fixed penalty (200) instead
of a projection-based value. \\
Dynamic-object handling &
None: no separate moving-object filter is applied beyond the depth/radius
and range-window checks above. This is disclosed as a limitation, not
assumed away. \\
Optimizer &
\texttt{scipy.optimize.minimize}, Powell method,
\texttt{maxiter=60}, \texttt{xtol=1e-4}, \texttt{ftol=1e-4}. \\
Search box $\mathcal{B}$ &
Rotation $\pm0.24$\,rad ($\approx13.7^\circ$) per axis, translation
$\pm0.45$\,m per axis -- comfortably above the severe tier's injected
range on both axes. \\
Initialization &
The canonical solve starts at $\Delta T=0$. Multistart/basin-consistency
solves add 5 further restarts drawn uniformly from the full search box (6
solves total per window); the mean pairwise geodesic rotation angle among
the 6 is the basin-spread confidence signal. \\
Legacy confidence heuristic &
$\text{improvement\_ratio}\times\text{point\_factor}$, clipped to
$[0,1]$, where \texttt{point\_factor}$=\min(1,\,n_{\text{points}}/500)$
and \texttt{improvement\_ratio} is the relative cost drop from start to
end of the optimization. \\
Camera &
\texttt{ring\_front\_center} for every AV2 experiment;
\texttt{CAM\_FRONT} for the nuScenes diagnostic. No other camera is used
anywhere in the paper. \\
Frame sampling &
Every 3rd LiDAR frame for the 27-log formal unit and the 60-log
exploratory pool; every frame for the learned corrector's 9-log evaluation
subset; consecutive frames from each 100-sweep system-audit bundle, with a
30-frame minimum and a 50-frame target. \\
Motion compensation &
Raw bundles preserve the LiDAR sweep and camera timestamps without using
recorded poses. The online mode estimates adjacent camera motion from
visual--LiDAR 3D--2D associations with RANSAC PnP and applies a fractional
$SE(3)$ transform for the timestamp gap. If estimation fails, the declared
fallback is used and the event is counted; recorded-pose compensation is
reported only as an exploratory dependency comparison. \\
System-audit guard &
Window length $W=3$, $M=2$ multistart restarts, $H=2$-window hysteresis,
and a 32-quantile threshold grid including a never-admit sentinel. The
locked final policy uses $\tau_b=\tau_a=0^\circ$ because no non-degenerate
candidate operating point met the conditional safety and coverage rules. \\
Random seeds &
A global seed of 1337 fixes all bootstraps, multistart draws, and system-audit
log ordering. Per-frame draws (point extraction, perturbation
sampling) instead use a deterministic seed derived from
\texttt{sha256(log\_id, frame\_idx, purpose)}, not a single shared stream,
so that adding or removing a log or a tier cannot change any other
frame's draw. \\
\end{longtable}
\twocolumn

\section{Protocol Details of the Bounded Visual--LiDAR Odometry Diagnostic}
\label{app:odometry}

This appendix gives the detailed protocol and audit record for the
visual--LiDAR odometry extension in the main text. The final result is
negative rather than a positive deployment claim: three development, fifteen
calibration, and fifteen final Argoverse~2 logs were assigned before
evaluation, and the final role was evaluated with the four-arm matrix. The
calibration role produced no non-degenerate safe operating point, so the
locked Guarded policy admitted no correction and matched Raw. The final
mixture nevertheless exposes the recursive-system risk: Always-correct has
substantially larger trajectory error than Raw, whereas Oracle provides a
non-deployable upper reference. Fourteen final logs reach the 50-frame target
and one is retained and marked short; all satisfy the 30-frame
minimum.

\subsection{Scope and Baseline-Selection Gate}

The extension evaluates a visual--LiDAR odometry front end in which
the camera--LiDAR extrinsic affects visual association, measurement
residuals, or state updates. It does not call a monocular visual SLAM
system (which does not consume this extrinsic) a valid test. Nor can a
LiDAR-only system such as LOAM \cite{zhang2014loam} be the sole primary
baseline: it is useful as a diagnostic localization reference, but it
cannot establish whether a camera--LiDAR correction is safe to admit.

The intended protocol prefers a maintained open-source visual--LiDAR or
LiDAR-visual-inertial implementation. The selected front end follows the
LIMO measurement pattern but is a dependency-light implementation for the
available Argoverse~2 bundles; it is not claimed to reproduce a complete ROS
mapping backend. Its result is therefore interpreted as a bounded,
project-native visual--LiDAR diagnostic contract, rather than as a benchmark claim
about every visual--LiDAR odometry system. The baseline was required to pass
the following compatibility checks:

\begin{enumerate}
\item Every required image, LiDAR, timestamp, calibration, and evaluation
field has an explicit source; no unavailable IMU input is silently emulated.
\item The candidate extrinsic enters point projection and visual--LiDAR
association before relative motion estimation, and the Oracle arm passes an
end-to-end health check.
\item A zero-injection smoke test runs end-to-end before any injected-drift
result is considered. It is a failure diagnostic, not a row that may be
discarded merely because the injected drift is zero.
\end{enumerate}

The selected AV2 development baseline is a dependency-light
LIMO-style implementation. It reads deterministic paired sensor bundles,
projects each unorganised point with the calibrated intrinsics and
camera$\leftarrow$ego transform, associates returns to LK feature tracks, and
solves the relative edge with RANSAC PnP. It intentionally avoids assumptions
about AV2's per-point scan-ring ordering and does not require ROS or IMU. The
three development logs produced valid three-tier, four-arm rollouts with
41, 75, and 97 usable frames; online motion estimation succeeded on every
adjacent edge in these development runs. The 15-log calibration and 15-log
final system audit are reported in the main text. The development smoke
passes the plumbing gate only; it does not by itself establish a positive
correction method.

LIMO's LiDAR-to-image depth association for tracked camera features
\cite{graeter2018limo} supplies the measurement pattern used by this
baseline. Visual-LiDAR Odometry and Mapping \cite{zhang2015visual}, LIO-SAM
\cite{shan2020liosam}, and FAST-LIO \cite{xu2021fastlio} clarify the
recursive-estimation context, but no claim is made that a complete mapping
backend has been reproduced.

\subsection{Correction Admission and Fallback}

For a fixed-extrinsic log session, a frozen correspondence/cost module
produces a candidate correction $\Delta\hat T_w$ on each window $w$ of
length $W$. Starting from the nominal initialization plus $M$ deterministic
random initializations, independently re-solve the same window and define
$b_w$ as the mean pairwise geodesic distance between solutions
(\emph{basin spread}). Let $a_w$ be the geodesic disagreement between the
candidate from the current window and the candidates in the preceding $L$
windows. A correction is admitted only if

\[
 b_w \leq \tau_b, \qquad a_w \leq \tau_a
\]

for $H$ consecutive windows. All $W,M,L,H,\tau_b,\tau_a$, random seeds,
and the acceptance-rate operating limit are selected on a calibration split
and then locked. On rejection, the front end uses the nominal injected
extrinsic $\hat T$; it does not insert a partial correction, alter a past
state, or use a score threshold selected on final logs. An admitted
correction is fixed for its declared segment, which makes state changes
attributable and prevents a single noisy window from repeatedly rewriting
the estimator.

\subsection{Split and Perturbation Contract}

Table~\ref{app:tab:split} fixes roles at log granularity. The existing
27-log formal set and 60-log exploratory set are evidence about the cost
surface, not a license to choose the system-audit guard and report its final
performance on the same data.

\begin{table}[h]
\centering
\small
\begin{tabular}{@{}p{0.23\linewidth}p{0.67\linewidth}@{}}
\toprule
Role & Exclusive use \\
\midrule
Development & Baseline, Oracle, zero-injection, and online-motion smoke
tests only; no reported threshold or final metric. \\
Calibration & Fix baseline configuration and $W,M,L,H,\tau_b,\tau_a$;
choose the guard operating point once. \\
Final & Run all arms and report intervals; never tune parameters or
replace a failed baseline. \\
External transfer & Deferred unless the Argoverse~2 system audit clears its
locked safety and benefit gates. \\
\bottomrule
\end{tabular}
\caption{Log-disjoint roles for the recursive experiment.}
\label{app:tab:split}
\end{table}

Each perturbation is sampled once per log/session and held fixed across
all frames and arms. The final report includes zero injection, mild, and
severe tiers, plus a tier-blind mixture. It may not infer an operating
decision from a confidence AUROC pooled across severity tiers.

\subsection{Four-Arm Rollout Matrix}

\begin{table}[h]
\centering
\small
\begin{tabular}{@{}p{0.25\linewidth}p{0.65\linewidth}@{}}
\toprule
Arm & Extrinsic policy \\
\midrule
Raw & Always use injected $\hat T$. \\
Oracle & Always use true $T$; upper reference and adapter health check. \\
Always-correct & Always inject $\Delta\hat T_w$; exposes recursive
amplification of a bad correction. \\
Guarded & Admit only temporal basin-consistent corrections; otherwise
fall back to Raw. \\
\bottomrule
\end{tabular}
\caption{All arms use identical timestamps, initial pose, data, and
odometry settings. Oracle is not a deployable method.}
\label{app:tab:arms}
\end{table}

The secondary ablations are (i) legacy cost-drop confidence, (ii) basin
spread only, and (iii) basin spread plus temporal agreement. The final
system audit reports all four paired arms rather than only accepted windows
or the always-correct arm.

\subsection{Admission Checklist}

The statistical unit is one log, with paired log bootstrap intervals;
frames within a trajectory are not independent replicates. Define the
pre-specified safety vector as endpoint ATE, fixed-horizon RPE,
tracking-loss/fallback rate, reprojection and association diagnostics, and
zero-injection acceptance rate.
If a front end does not expose a safety field, the field is reported as
unavailable and the corresponding safety gate cannot be claimed to pass.
The non-inferiority margins and the fixed RPE horizons are selected on
calibration data before final runs. Table~\ref{app:tab:gates23} states the
decision rule.

\begin{table}[h]
\centering
\small
\begin{tabular}{@{}p{0.45\linewidth}p{0.45\linewidth}@{}}
\toprule
Criterion & Required outcome \\
\midrule
Benefit & Guarded improves Raw on ATE or RPE in the tier-blind final
mixture; paired-log 95\% CI excludes zero. \\
Safety & Guarded is non-inferior to Raw on every predeclared safety
metric, including zero injection. \\
Mechanism & Always-correct versus Guarded is reported. If it is worse,
this supports recursive-risk mitigation; if not, no necessity claim. \\
Failure handling & Any failed criterion is a negative result. Parameters,
thresholds, and final logs are not recycled for a new attempt. \\
\bottomrule
\end{tabular}
\caption{Admission gates for the visual--LiDAR odometry extension.}
\label{app:tab:gates23}
\end{table}

\subsection{Observed System-Audit Result}

The calibration role used $W=3$, $M=2$, $H=2$, and a 32-quantile grid for
basin-spread and temporal-agreement thresholds, including an explicit
never-admit sentinel. It produced 2,748 candidate observations over 15
logs. Harmful-admission risk was computed conditionally among admitted
zero-injection candidates; an empty admitted set receives an upper bound of
one, so reject-all cannot satisfy the safety rule. No candidate operating
point met both the conditional safety constraint and the requirements for
coverage across logs, injected-tier admission, and at least one helpful
admission. The locked policy is therefore the negative-control
never-admit policy with $\tau_b=\tau_a=0^\circ$.

The final tier-blind mixture contains 15 logs, 45 tier rollouts, and 180 arm
records. Fourteen logs reach the 50-frame target and one is marked
short; all reach the 30-frame minimum. The absolute ATE means are
$7.458$\,m (Raw), $4.520$\,m (Oracle), $20.474$\,m (Always-correct), and
$7.458$\,m (Guarded). The corresponding 10-frame joint RPE means are
$3.215$, $2.240$, $7.246$, and $3.215$, respectively. The Guarded--Raw ATE
difference is $0.000$\,m with 95\% CI $[0.000,0.000]$, while
Always-correct--Raw is $-13.017$\,m with 95\% CI
$[-18.639,-7.729]$. Across Raw and Guarded, online motion estimation
succeeded on 3,291 of 3,297 adjacent edges; the six remaining edges used
the declared fallback. Tracking loss, reprojection error, and association
coverage are reported in the main system table. The negative-control result
does not certify tracking safety or deployment benefit.

\subsection{Additional Guard Diagnostics}

The final guard uses only quantities available before a correction is applied:
basin spread across deterministic restarts and temporal agreement with recent
candidate estimates. Earlier diagnostics considered alternative signals,
including cost-drop confidence and cross-modal consistency. They are not used
to select the final operating point. The final calibration instead performs a
single pre-specified grid search over the observable basin and temporal
features, with the conditional zero-injection safety rule and the
non-degeneracy requirements described above.

This separation is important for interpreting the negative result. A ranking
statistic can identify a subset of corrections that is more promising on a
retrospective sample without supplying a safe threshold at nonzero coverage.
The final calibration did not find such a threshold. The never-admit policy
is consequently reported as a negative control, not as evidence that a
different future guard could not succeed.

\begin{table*}[t]
\centering
\scriptsize
\begin{tabular}{@{}llrrrr@{}}
\toprule
Role & $s_t$ (m) & Observed & Null mean & Observed--null [95\% CI] & $p<0.05$ logs \\
\midrule
Formal 27-log & 0.1 & 6.523 & 8.944 & $-2.421$ [$-3.595,-1.225$] & 11/27 \\
Formal 27-log & 1.0 & 4.990 & 6.314 & $-1.324$ [$-2.431,-0.286$] & 7/27 \\
Formal 27-log & 10.0 & 4.965 & 6.259 & $-1.294$ [$-2.412,-0.253$] & 7/27 \\
Exploratory 60-log & 0.1 & 8.069 & 10.164 & $-2.095$ [$-3.176,-1.126$] & 21/60 \\
Exploratory 60-log & 1.0 & 5.734 & 7.540 & $-1.806$ [$-3.032,-0.741$] & 22/60 \\
Exploratory 60-log & 10.0 & 5.686 & 7.482 & $-1.796$ [$-3.028,-0.728$] & 20/60 \\
\bottomrule
\end{tabular}
\caption{Pre-specified sensitivity of the shared-attractor composite to the
translation scale. Rotation axes, null direction seed (1337), 20,000 null
draws per log, and 10,000 log-level bootstrap draws are held fixed across
scales. The qualitative sign is stable, while the composite magnitude and
tail-log count are scale-dependent.}
\label{tab:scale-sensitivity}
\end{table*}

\subsection{Metric Definitions and Long-Horizon Reporting}

For a trajectory with estimated poses $\{\hat{T}_i\}$ and reference poses
$\{T_i\}$, absolute trajectory error is reported as the position RMSE after
the declared trajectory alignment. Fixed-horizon relative pose error at
horizon $h$ is computed from
\[
 E_i^{(h)} =
 \left(T_i^{-1}T_{i+h}\right)^{-1}
 \left(\hat{T}_i^{-1}\hat{T}_{i+h}\right),
\]
using the rotational geodesic angle and translation norm of
$E_i^{(h)}$. The primary fixed-horizon comparison uses $h=10$ frames;
$h\in\{1,5,25\}$ is retained as secondary trajectory evidence. The joint
score in the main table is the declared composite
\[
 d_{\mathrm{joint}} =
 \sqrt{\theta^2 + \left(\frac{\|t\|}{1\,\mathrm{m}}\right)^2},
\]
where $\theta$ is in radians. It is a reporting composite and is not claimed
to be an invariant metric on $SE(3)$. The primary scale is $1\,\mathrm{m}$;
the pre-specified translation-scale sensitivity is reported below.

The sensitivity audit does not alter the primary $1\,\mathrm{m}$ analysis or
any admission threshold. It bounds the interpretation: the shared-attractor
pattern is a reproducible observational co-occurrence under all three
declared scales, not a scale-free causal or geometric certificate.

Reprojection RMSE is computed for associated LiDAR returns projected into
the camera image, while association coverage is the fraction of tracked
visual features with a valid LiDAR association. Tracking loss and online
motion-estimation fallback are counted per adjacent edge. Because no
high-definition map archive was available for these bundles, reprojection and
association are reported as map-independent consistency proxies rather than
as map-consistency scores. All trajectory summaries average within each log
before paired bootstrap resampling of logs; frames and edges are not treated
as independent replicates.

\subsection{Scope of the Negative Result}

The final audit demonstrates that the online consumer, full six-DoF
trajectory metrics, fixed-horizon evaluation, reprojection and association
diagnostics, and long-horizon data contract can be executed under a
log-disjoint protocol. It also shows that Always-correct can amplify a wrong
correction substantially, while the locked Guarded policy remains identical
to Raw because calibration found no non-degenerate admitted set. These
observations support the stated audit conclusion, but they do not establish
deployment safety, map quality, or a positive recursive correction method.
The deferred cross-domain transfer and the absence of a full mapping backend
remain explicit scope limitations.